\pdfoutput=1

\documentclass[11pt]{article}

\usepackage[]{naacl2021}
\usepackage{times}
\usepackage{latexsym}

\usepackage[T1]{fontenc}

\usepackage[utf8]{inputenc}

\usepackage{microtype}

\usepackage{multirow}
\usepackage{amsmath}
\usepackage{amsfonts}
\usepackage{colortbl}
\usepackage{graphicx, subfigure}

\usepackage{xcolor,colortbl}

\newcommand{\modelname}{\textsc{Entailment}}
\newcommand{\hybrid}{\textsc{Hybrid}}
\newcommand{\intentdata}{\textsc{IFS-Intent}}
\newcommand{\relationdata}{\textsc{IFS-Relation}}

\title{Incremental Few-shot Text Classification with Multi-round New Classes:\\Formulation, Dataset and System}

\author{Congying Xia$^{1}$\thanks{~ Indicates Equal Contribution.}, Wenpeng Yin${^2\footnotemark[1]}$, Yihao Feng$^{3}$, Philip Yu$^{1}$\\
{$^1$University of Illinois at Chicago, Chicago, IL, USA} \\
{$^2$Salesforce Research, Palo Alto, CA, USA} \\
{$^3$University of Texas at Austin, Austin, TX, USA} \\
 {\tt \{cxia8, psyu\}@uic.edu},
 {\tt wyin@salesforce.com},
 {\tt yihao@cs.utexas.edu}
 }

\begin{document}
\maketitle
\begin{abstract}

Text classification is usually studied by labeling natural language texts with relevant categories from a predefined set. In the real world, new classes might keep challenging the existing system with limited labeled data. The system should be intelligent enough to recognize upcoming new classes with a few examples. In this work, we define a new task in the NLP domain, \textit{incremental few-shot text classification}, where the system incrementally handles multiple rounds of new classes. For each round, there is a batch of new classes with a few labeled examples per class. Two major challenges exist in this new task: (i) For the learning process, the system should incrementally learn new classes round by round without re-training on the examples of preceding classes; (ii) For the performance, the system should perform well on new classes without much loss on preceding classes. In addition to formulating the new task, we also release two benchmark datasets \footnote{Code \& Data are available at \url{https://github.com/congyingxia/IncrementalFSTC}.} 
in the incremental few-shot setting: intent classification and relation classification. Moreover, we propose two entailment approaches, \modelname\enspace and \hybrid, which show promise for solving this novel problem.
\end{abstract}

\section{Introduction}
Text classification has achieved great success in the past decades with the development of deep learning techniques \cite{kowsari2019text, li2020survey}. However, decent performance highly relies on the availability of large-scale task-specific training data. Recently, few-shot text classification \cite{yu2018diverse, geng2019induction, xia2020composed} has attracted increasing attention from the NLP community since it is unlikely to have large-scale labeled data for new classes in the real world. 

Typically, few-shot text classification is formulated like this: the system first sees a set of base classes $C_b$ that have a large number of labeled examples,
then a group of new classes $C_n$ is provided with $k$ examples per class.
For a testing instance, the system is required to search for its label in the space of $C_b\cup C_n$ or merely $C_n$. However, this setting might not suitable for real scenarios. First, base classes with rich annotations might not be available at the beginning. It happens whenever you want to build a system from scratch. Second, take the bank's customer service system as an example, queries with new intents are continuously appearing (e.g., by a sequence of rounds) without enough labeled data. The system should be able to keep learning and recognizing new intents round by round. For each query, the system needs to pick up the most appropriate intent in the incrementally increasing label space or return ``none of them''.

In this work,
we propose a more realistic and challenging task in the low resource scenarios: \textit{incremental few-shot text classification}. In this new task, the system is provided with $m$ rounds of new classes (i.e., $C^1_n$, $C^2_n$, $\cdots$, $C^m_n$) without any base classes that have enough annotations. For each round, there are a group of new classes, $C^i_n$ ($i=1, \cdots, m$), and each class has $k$ labeled examples ($k$ is in the range of [1, 5] and varies for different classes). During testing, the system is required to either select the best class from $C^1_n \cup C^1_n\cdots \cup C^m_n$ or output ``none of them'' which means no existing class applies to the input. As far as we know, this is the first work that studies incremental few-shot learning without base classes. All previous few-shot learning models \cite{DBLPSnellSZ17, DBLPGidarisK18, yin2020universal, xia2020cg, nguyen2020semantic} fail to solve this problem since they relied on the large-scale labeled data of base classes to train a robust system. To provide a complete vision about incremental few-shot text classification, we also conduct experiments with additional base classes to compare with these baselines.

To evaluate the performance for different models, we build two benchmark datasets for this new problem. One is intent detection that aims at understanding the intents under user queries \cite{liu2016attention, xia2018zero}. This benchmark simulates a task like a bank's customer service as we mentioned. The other is relation classification which needs to determine the correct relation between two entities in a given sentence \cite{zeng2014relation}. In reality, the relation types might be unlimited. For example, there are fine-grained relations or implicit relations that need entailment. In open-domain or open-form relation tasks, there always exists the problem of lack of annotations. 

Another important feature of our benchmark datasets is that we do not provide dev sets. Existing systems are commonly evaluated on the dev set to choose the best training model. We claim that in real-world (incremental) few-shot applications, we cannot expect extra labeled data other than the $k$ examples. This is in line with the observation in \citet{DBLP07676}. If a system has to rely on the dev set to find the best parameters, it is not suitable for the incremental few-shot setting.



Furthermore, we propose a novel approach, \modelname, to solve this new problem. \modelname\enspace models the text classification problem in a textual entailment \cite{dagan2013recognizing} framework. To figure out if an input $x$ belongs to a class $y$, \modelname\enspace tries to infer the truth value of $y$ (i.e., a hypothesis), given the $x$ (i.e, the premise). The main benefit of this formulation is that the system learns this task not only from label-specific examples, but more importantly, from the large-scale entailment datasets. In other words, we make use of indirect supervision from textual entailment datasets to address the target few-shot task. 

In summary, our contribution lies in three aspects. 1) We propose a new task named Incremental Few shot Text Classification with multi-round new classes. This task is more challenging and realistic for low resource scenarios. 2) We create and release two benchmark datasets to evaluate the performance of this new task. 3) We propose two novel models, \modelname\enspace and \hybrid, to solve this novel problem. Extensive experiments on these two datasets show the effectiveness of our proposed models.

\section{Related Work}
\label{sec:relatedwork}
\paragraph{Incremental few-shot learning.} As far as we know, there is no prior work in the NLP domain that studies incremental few-shot text classification. In this section, we mainly introduce some work in the computer vision domain. These works only assume that a single round of new classes $C_n$ is appended to the base classes $C_b$. Generally, they will learn class representations for classification.
Different approaches differ in the way of representing base classes and new classes. Hereafter, we use $W_b$ and $W_n$ as the representations for $C_b$ and $C_n$, respectively.


\newcite{DBLPSnellSZ17} proposes the Prototypical Network, in which both $W_b$ and $W_n$ are stored as the average embedding of the few-shot support images for a certain class.
Although Prototypical Network was not designed for incremental few-shot learning, it can be easily adapted to the incremental setting by providing the representations for all the classes. It trains a nearest neighbor algorithm on the base classes and tests directly on the union of base and new classes. \newcite{DBLPQiBL18} proposes an ``imprinting'' mechanism: the base representations $W_b$ are learned through supervised pre-training (e.g., the weight matrix in a softmax classifier), and $W_n$ are computed using the averaged representations like Prototypical Network.


In \citet{DBLPGidarisK18}, the base representations $W_b$ are learned through supervised pre-training. The representation of the $i^{th}$ novel class ($W_{n,i}$) comes from two origins: (i) the prototypical averaging, $w_{avg}$; (ii) attention-weighted sum over base representations: $w_{att}$. Namely,  $W_{n,i}=\phi_{avg}\odot w_{avg}+\phi_{att}\odot w_{att}$, where $\phi_{avg}$ and $\phi_{att}$ are learnable weight vectors. In the few-shot training stage, the original base classes $C_b$ are split into ``new base classes'' and ``fake novel classes'' for each episode. In testing, the representations of novel classes, $W_n$, are constructed based on the $k$ examples and $W_b$.



In \citet{DBLPRenLFZ19}, both $W_b$ and $W_n$ are learned through supervised training:  $W_b$ are classifier parameters pre-trained on base classes, $W_n$ are classifier parameters learned in new classes.
During the training, the support set and the query set are constructed differently for new classes.
The support set consists of examples only from new classes; the query set contains examples from both new classes and base classes (because the training goal is to maximize the performance of all classes). The training in this literature has two phases. The first phase is few-shot episode training which learns $W_n$, the second phase (called meta-learning training) optimizes the performance on the query set and regularizes the representations for new classes.


To summarize, compared with \citet{DBLPSnellSZ17} and \citet{DBLPQiBL18}, both \citet{DBLPGidarisK18} and \citet{DBLPRenLFZ19} build connections between the representations of base classes and the new classes.
However, these methods cannot be directly applied to our problem for the following reasons. (i) Despite the claims in some literature that they are dealing with incremental or dynamic few-shot problems, they only considered a single round of new classes \cite{DBLPQiBL18,DBLPGidarisK18,DBLPRenLFZ19}. It is unclear if the system can keep the performance when multi-round new classes are considered.
(ii) During the training for the new classes, they often rely on extra labeled data other than the $k$ examples, such as the query set in \citet{DBLPRenLFZ19}.
(iii) Different from their setting, we have an extra label ``none-of-them'' in incremental few-shot text classification. It's not guaranteed that the input, such as the customer's utterance, always falls into the range of seen labels.

\paragraph{Using textual entailment for text classification.}
\citet{zhangdiscriminative} is a state-of-the-art paper for few-shot text classification. 
They propose a clustering-based classifier named discriminative nearest neighbor classification (DNNC). DNNC compares whether two examples are in the same class or not.
A matching model S($x_i$, $x_j$) is trained as a binary classifier, such that S($x_i$, $x_j$) is close to 1.0 if $x_i$ and $x_j$ belong to the same class, otherwise close to 0.0. Thus, their model can be pre-trained with a large-scale textual entailment dataset. Given a test query $x$, they compare the test query with all the previous examples. The final prediction is made by searching the nearest neighbor which has the highest matching score S($x$, $x_i$) with the query example. 
Their computation cost is high due to the comparision between all the utterance pairs.

Moreover, comparing whether two examples are in the same class is different from textual entailment. In textual entailment, a person reads a premise to infer that the hypothesis is true or not. The fact that two examples are in the same class does not mean they can entail each other.
Thus, they cannot fully utilize the pre-trained entailment model.
Instead, our proposed model, \modelname, entails the label with a given utterance, which is much more efficient and maximizes the utilization of the pre-trained entailment model. 

\citet{DBLPYinHR19} is another work that utilizes textual entailment for zero-shot text classification. They convert the zero-shot text classification as a problem of filling a label for a hypothesis. For example, they combine ``emotion'' labels with the question “this text expresses ?”, and ask the model if this hypothesis is true, given the text. This work more focuses on zero-shot learning and they need to propose different questions for different labels.

\section{Problem Formulation}\label{sec:probdefinitino}
In this section, we give a formal description of the problem ``incremental few-shot text classification'' without base classes. Furthermore, we extend the problem with additional base classes.

\paragraph{Training data.} In the incremental few-shot text classification setting, the system is provided with $m$ rounds of new classes sequentially: \{$C_n^1, \cdots, C_n^m$\}. Each round $C_n^i$ has $h$ new classes, namely $C_n^i=\{C^i_{n,1}, \cdots, C^i_{n,h}\}$. Each new class only has $k$ examples ($k\in [1,5]$). The value of $k$ is not fixed and varies for different new classes in the same round, i.e., $k_{C^i_{n,s}}\neq k_{C^i_{n,t}}$, where $s, t \in [1, ..., h]$. For the setting with additional base classes, the system can access a set of base classes $C_b=\{C_{b,1}, C_{b,2},\cdots, C_{b,g}\}$. All the base classes $C_{b}$ have enough labeled examples for training. 

We create the multi-round setting to mimic the real-world scenario where there is a sequence of new classes coming to the system. 
Since we can only collect a handful of examples for the upcoming classes and the number of examples cannot be guaranteed, we set $k\in [1,5]$ and allow the flexibility that $k_{C^i_{n,s}}\neq k_{C^i_{n,t}}$ in each round.

\paragraph{Development data.} 
In the incremental few-shot setting, there are only $k$ examples available for each new class. Thus, our formulation does not provide any development set to help select the best model. It is recommended to select hyper-parameters based on experience or related tasks. In the experiments, we choose hyper-parameters like batch size based on the suggestions by Huggingface\footnote{\url{https://github.com/huggingface/transformers}} and other papers like \citet{devlin2018bert} and \citet{zhangdiscriminative}.


\paragraph{Testing data.}
To evaluate the system, the test data consists of examples across all the classes. For the setting without base classes, the potential label space is $C_{n}^1 \cup \cdots \cup C_{n}^m \cdots \cup C_o$. For the setting with additional base classes, we search among all the classes in $C_b\cup C_{n}^1 \cup \cdots \cup C_{n}^m \cdots \cup C_o$. $C_o$ is an extra out-of-distribution (OOD) class that consists of examples falling outside of all the seen classes. It gives us a chance to check the system's ability to detect instances that reject all the known classes. This is crucial for an open-set problem like incremental learning since there are always examples from upcoming classes that do not belong to any existing class.

\paragraph{Requirements.}
(i) For the training of $i^{th}$ round $C_{n}^i$, the system can only access the newly added few-shot examples and label names in this round. The system is not allowed to re-train on the (full or partial) examples of preceding classes.
(ii) For the evaluation, we care about the performance in different types of classes, including base classes, different rounds of new classes, and OOD classes in $C_o$. We expect a system that can continuously recognize new classes with few-shot examples. In the meantime, the performance of preceding classes should be maintained.
A system showing severer catastrophic forgetting is less preferred.

\section{Our Model: \modelname}
Our approach \modelname\enspace casts the text classification problem into textual entailment: the input text acts as a premise, the class name, such as ``open a bank account'' in intent detection
, acts as a hypothesis. Then the question that if the input belongs to a class is equivalent to ask if the hypothesis is true given the premise. There are two benefits of transforming the text classification problem to entailment. 
First, we can make use of indirect supervision from a large-scale entailment dataset \cite{williams2018broad} to benefit the few-shot settings.
Second, this enables us to utilize the few-shot examples as well as the information of the class names. Typical text classification approaches treat classes as indices. In fact, class names usually contain informative signals.

\paragraph{Entailment pairs.} 
To transfer the text classification problem into textual entailment, we construct positive and negative entailment pairs for the training.
Positive entailment pairs ($x_i$, $y_i$) are constructed with utterance $x_i$ and its gold label name $y_i$, where $y_i\in C_b$ for base classes and $y_i\in C_n^i$ for new classes. Negative entailment pairs consist of ($x_i$, $y_j$), where $y_j$ is an incorrect label in the current round. For base classes, $y_j\in C_b$ but $y_j\neq y_i$; for new classes, $y_j\in C_n^i$ but $y_j \neq y_i$.

For each entailment pair ($x, y$) whether it is positive or negative, we concatenate its utterance $x$ with the label $y$ and fed it into the RoBERTa \cite{liu2019roberta} encoder.
Given an utterance $x = (X_1, X_2, ..., X_{T_2})$ with $T_1$ words and a label $y = (Y_1, Y_2, ..., Y_{T_{2}}) $ with $T_2$ words, we add a special start-of-sequence ([CLS]) token at the beginning of the input and a special end-of-sequence ([SEP]) token at the end of each sentence. The whole input is ([CLS], $X_1$, $X_2$, ..., $X_{T_1}$, [SEP], $Y_1$, $Y_2$, ...,  $Y_{T_{2}}$, [SEP]). We use the [CLS] embedding output from the RoBERTa encoder with a fully connected layer for binary textual entailment:
\begin{align}
    e =& \text{RoBERTa}(x, y),\\
    p =& \text{softmax}(We + z),
\end{align}
where $h \in \mathbb{R}^{d}$ is the embedding for the [CLS] token, $W \in \mathbb{R}^{2 \times d}$ and $z \in \mathbb{R}^2$ are parameters.

Compared to \citet{zhangdiscriminative}, they discriminate whether two utterances ($x_i, x_j$) are in the same class or not.
($x_i, x_j$) is a positive pair if they belong to the same class, otherwise, it is a negative pair. To explore the potential of different combinations, we also propose a hybrid entailment model, \hybrid, that uses both (utterance, label) pairs ($x_i, y_i$) and (utterance, utterance) pairs ($x_i, x_j$). In other words, we train \hybrid\enspace with pairs from both \modelname\enspace and DNNC \cite{zhangdiscriminative}. In round $C_n^i$ which contains $h$ new classes and $k$ examples for each class, \modelname\enspace generates $h*k$ positive entailment pairs and $(h-1)*h*k$ negative entailment pairs, while DNNC generates $h*k*(k-1)$ positive pairs and $h*(h-1)*k^2$ negative pairs. \hybrid\enspace utilizes pairs from both models. For simplicity, we use the same $k$ value for all new classes here; in real datasets, different new classes may have different numbers of few-shot examples. In that case, the number of generated pairs will change accordingly.

\paragraph{Training strategy.} 

Both \modelname\enspace and \hybrid\enspace are binary classification models that can utilize indirect supervision from textual entailment. Firstly, we pre-train these models with a large-scale entailment dataset \cite{williams2018broad}. For each round, models are fine-tuned on the new classes in $C_n^i$. For the setting with additional base classes, we fine-tune the models on base classes first. Then we continuously fine-tune the models on new classes.

\paragraph{Inference strategy.} After the training, we use the model to infer the class for a test input. For each input utterance, we generate entailment pairs by accompanying the utterance with all classes except $C_o$. Each pair will get a score $\lambda \in [0, 1]$ indicating whether this input belongs to the particular class or not. $\lambda > 0.5$ indicates ``YES'', ``No'' otherwise. If there is at least one class labeled with ``YES'',  the class with the maximal $\lambda$ score is returned; otherwise, the system returns $C_o$. We choose the threshold as 0.5 because entailment recognition is a binary classification problem.

Next, we compare our model with some related systems that can be potentially applied to the incremental few-shot text classification.

\paragraph{\modelname\enspace vs. Prototypical Network.} Prototypical Network \cite{DBLPSnellSZ17} tries to solve few-shot target tasks given a collection of training tasks.
The few-shot learning problem solved in Prototypical network is slightly different from our incremental few-shot setting. In Prototypical Network, the label space for target tasks only contains the new classes. However, in the incremental few-shot setting, the target label space is continuously increasing by adding new classes. Due to this essential distinction, applying Prototypical Network to incremental few-shot are very likely to have performance drop on base classes when fine-tuning on new classes.


\paragraph{\modelname\enspace vs. Incremental few-shot approaches in computer vision.} In Related Work, we introduced some typical approaches in computer vision that deal with the incremental few-shot problem. Those methods consistently try to learn representations for classes and examples separately (i.e,, the $W_b$ and $W_n$ in Section \ref{sec:relatedwork}). In our model, there are no individual representation vectors for classes or examples. 
Instead, the model learns an overall representation vector for the whole (input, class) pair. Our solution enables the learning of the input and the class to interact with each other, which has widely demonstrated its superiority in modeling the relations of two elements \cite{DBLP12808,zhangdiscriminative}.

In addition, the approaches in computer vision mostly rely on large-scale labeled data for base classes to train a robust system. We would argue that the base classes with rich annotations may not be available in real-world applications. Our system which can be pre-trained with entailment dataset, instead, does not rely on base classes. This makes our system more applicable to various scenarios.

\begin{table}[t]
\centering
\resizebox{\linewidth}{!}{
\begin{tabular}{l|c|c|c||c|c|c}
& \multicolumn{3}{c||}{\intentdata}&  \multicolumn{3}{c}{\relationdata} \\
  & \#class & \#train  & \#test & \#class & \#train  & \#test \\\hline
$C_b$ & 20 & 2088  & 800 & 10 & 5000 & 400\\
$C_n^1$ & 10 &  30   & 400 & 10 & 30 & 400 \\
$C_n^2$ &  10 & 30   & 400 & 10 & 30 & 400\\
$C_n^3$ &  10 &  30   & 400 & 10 & 30 & 400 \\
$C_n^4$ &  10 &  30   & 400 & 10 & 30 & 400 \\
$C_n^5$ &  10 &  30   & 400 & 10 & 30 &400 \\
$C_o$ & 7 & --   & 280 & 10 &  - & 400 \\
\end{tabular}
}
\vspace{-0.05in}
\caption{Statistics of two datasets: {\intentdata} and {\relationdata}. $C_b$: base classes; \{$C_n^1$, $\cdots$, $C_n^5$\}: five rounds of new classes; $C_o$: OOD classes. Note that $C_o$ is never used for training.}\label{tab:intentstatistic}
\end{table}

\begin{table*}[t]
\setlength{\tabcolsep}{4.5pt}
\small
  \centering
  \begin{tabular}{ll|c|c|c|c|c|c}
 &  & $C_n^1$ & $C_n^2$ & $C_n^3$ & $C_n^4$ & $C_n^5$ & $C_o$\\\hline\hline 

 \multirow{3}{*}{$C_n^1$} 
 &  DNNC & 55.50$\pm$2.27 & & & & & 72.29$\pm$0.20\\
 & \modelname &65.17$\pm$1.36 & & & & &75.43$\pm$0.41 \\
  &  \hybrid & \textbf{70.08$\pm$0.77} & & & & & \textbf{78.25$\pm$0.19}\\\hline

 \multirow{3}{*}{$C_n^2$} &  DNNC &64.58$\pm$0.42 &77.75$\pm$1.08 & & & & 61.72$\pm$0.90\\
 &   \modelname &64.08$\pm$2.04 &76.33$\pm$1.01 & & & &\textbf{64.68$\pm$0.71}\\
  & \hybrid  & \textbf{74.25$\pm$1.34} &\textbf{86.67$\pm$1.01} & & & &64.39$\pm$0.27 \\\hline
 
  \multirow{3}{*}{$C_n^3$} &  DNNC & 65.25$\pm$1.67 &79.58$\pm$1.50 &64.67$\pm$1.93 & & & 50.25$\pm$0.52\\
  &   \modelname  &\textbf{75.50$\pm$1.63} &83.83$\pm$0.62 &75.25$\pm$1.24 & & &\textbf{56.56$\pm$2.43} \\
  & \hybrid & 74.25$\pm$1.08 & \textbf{85.92$\pm$1.05}& \textbf{76.58$\pm$1.05}& & &53.09$\pm$1.73 \\\hline
  
   \multirow{3}{*}{$C_n^4$}& DNNC &66.75$\pm$0.54 &79.08$\pm$0.51 & 60.50$\pm$2.35& 62.25$\pm$1.08& &  42.56$\pm$0.76\\
   &   \modelname &68.33$\pm$1.16 &72.67$\pm$0.77 &68.58$\pm$1.90 & 69.50$\pm$1.34& &\textbf{53.92$\pm$0.75}\\
  
  &  \hybrid & \textbf{73.75$\pm$1.41} &\textbf{85.50$\pm$1.06} & \textbf{71.67$\pm$1.53}& \textbf{75.83$\pm$2.44}& & 52.75$\pm$0.63\\\hline
  
   \multirow{3}{*}{$C_n^5$} &  DNNC &65.33$\pm$0.62 & 76.75$\pm$1.59& 62.83$\pm$3.17&59.75$\pm$2.83 &57.25$\pm$2.32 & 36.66$\pm$1.07\\
   &  \modelname &67.58$\pm$0.82 & 73.50$\pm$1.24& 67.83$\pm$0.47& 71.83$\pm$0.66& \textbf{73.75$\pm$0.74}&\textbf{50.95$\pm$0.68} \\
  &  \hybrid &\textbf{70.75$\pm$1.27} &\textbf{82.50$\pm$1.27} & \textbf{72.42$\pm$0.96}&\textbf{76.67$\pm$1.05} & 71.00$\pm$0.41& 47.05$\pm$1.60\\\hline
\end{tabular}
\caption{System performance without base classes on the benchmark \intentdata. Horizontal direction: different groups of testing classes (base classes $C_b$, five rounds of novel classes ($C_n^1, \cdots, C_n^5$) and the OOD classes $C_o$); vertical direction: timeline of incremental learning over new rounds of novel classes.  Numbers are averaged over results of three random seeds.}\label{tab:intentresults_withoutbase}
\end{table*}

\begin{table*}[t]
\setlength{\tabcolsep}{4.5pt}
\small
  \centering
  \begin{tabular}{ll|c|c|c|c|c|c}
 &  & $C_n^1$ & $C_n^2$ & $C_n^3$ & $C_n^4$ & $C_n^5$ & $C_o$\\\hline\hline 

 \multirow{2}{*}{$C_n^1$} 
 &  DNNC & 12.17$\pm$0.88 & & & & & 28.89$\pm$13.39 \\
 & \modelname &\textbf{67.17$\pm$1.20} & & & & &\textbf{82.03$\pm$6.36}\\\hline

 \multirow{2}{*}{$C_n^2$} &  DNNC &6.47$\pm$1.02 &5.28$\pm$0.75 & & & &73.97$\pm$4.83  \\
 &   \modelname &\textbf{51.67$\pm$5.02} &\textbf{53.00$\pm$3.01} & & & &\textbf{81.61$\pm$4.71}\\\hline
 
  \multirow{2}{*}{$C_n^3$} &  DNNC & 3.5$\pm$1.26 & 3.83$\pm$0.51& 2.28$\pm$1.09& & &   74.56$\pm$5.56\\
  &   \modelname  &\textbf{52.83$\pm$0.66} &\textbf{36.50$\pm$6.82} &\textbf{56.33$\pm$3.50} & & &\textbf{44.06$\pm$20.17} \\\hline
  
   \multirow{2}{*}{$C_n^4$}& DNNC & 1.67$\pm$0.89 & 2.39$\pm$0.45& 2.64$\pm$0.92 & 4.31$\pm$0.41 & &43.1$\pm$13.97 \\
   &   \modelname &\textbf{40.58$\pm$3.71} &\textbf{42.17$\pm$5.87} &\textbf{47.17$\pm$7.74} & \textbf{34.92$\pm$4.09}& &\textbf{78.57$\pm$2.15}\\\hline
  
   \multirow{2}{*}{$C_n^5$} &  DNNC & 1.47$\pm$0.39 &2.44$\pm$0.7 & 2.64$\pm$1.44 &1.08$\pm$1.12& 2.42$\pm$0.35 & 20.03$\pm$7.29 \\
   &  \modelname &\textbf{32.08$\pm$7.00} & \textbf{34.75$\pm$2.16}&\textbf{37.67$\pm$7.29}& \textbf{24.58$\pm$2.63}& \textbf{22.50$\pm$3.18}&\textbf{22.29$\pm$14.49}\\ \hline
\end{tabular}
\caption{System performance without base classes on the benchmark \relationdata.
}\label{tab:relationresults_withoutbase}
\end{table*}

\section{Experiments}

\subsection{Datasets}

\paragraph{\intentdata.} This is our benchmark for incremental few-shot intent detection. \intentdata\enspace is converted from BANKING77\footnote{\url{https://github.com/PolyAI-LDN/task-specific-datasets}} \cite{casanueva2020efficient}, which is a single-domain intent
detection dataset comprising 13,083 annotated
examples over 77 intents (average: 170 examples per intent). Each intent class is described by a short name, such as ``get physical card'', ``lost or stolen card'', etc. We randomly split the 77 intents into a base group (i.e., $C_b$, 20 base classes), 5 rounds of new intents (i.e., \{$C_n^1$, $\cdots$, $C_n^5$\}, each round has 10 new classes), and a group of out-of-distribution intents (i.e., $C_o$, 7 ood classes). 


\paragraph{\relationdata.} This is the benchmark for incremental few-shot relation classification. \relationdata\enspace is converted from FewRel\footnote{\url{https://github.com/thunlp/FewRel}} \cite{han2018fewrel}, which is a large-scale relation classification dataset. FewRel contains relations from different domains, including Wikipedia \cite{vrandevcic2014wikidata}, SemEval-2010 \cite{hendrickx2019semeval} and Pubmed\footnote{\url{https://www.ncbi.nlm.nih.gov/pubmed/}}. For classes in $C_b, C_n^1, C_n^2, C_n^3$, $C_n^4$, we randomly sample 10 classes from Wikipedia. Classes in $C_n^5$ come from SemEval-2010 and classes in $C_o$ come from Pubmed. 

Details for two datasets are reported in Table \ref{tab:intentstatistic}. For both benchmarks, we first split the classes into different rounds according to the setting illustrated in Table \ref{tab:intentstatistic}. Then we split the train/test examples provided by the original dataset into different rounds according to the split classes. For the new classes in each round, we randomly split 10 new classes into 5 groups (each with 2 classes) and intentionally let the 5 groups have different sizes of k-shot examples ($k\in [1,5]$).

\subsection{Experimental setting}
\paragraph{Baselines.} 
Since this is the first work that studies the incremental few-shot text classification problem, there is no prior system that deals with exactly the same task. In the setting without base classes, most few-shot learning models didn't work. We compare our proposed model \modelname\enspace with another work \cite{zhangdiscriminative} which also solves text classification as a textual entailment problem and use large-scale entailment datasets for pre-training. Together, their hybrid model, \hybrid, is also compared. In the setting with additional base classes, we further compare two few-shot learning models \cite{DBLPSnellSZ17, DBLPGidarisK18} adapted from the computer vision field. For these two baselines, we replace their encoders with RoBERTa to fit into the text classification task.

\textbullet\enspace \textbf{DNNC}. \newcite{zhangdiscriminative} proposed a discriminate nearest neighbor classifier. They decide whether two utterances are in the same class or not and make predictions by assigning the label of the nearest neighbor among all the examples.

\textbullet\enspace \textbf{Prototypical Network} \cite{DBLPSnellSZ17}. We train the Prototypical Network on base classes with the episode training method. For each round $C_n^i$, representations for new classes are calculated as the average embedding of $k$-shot examples. Given a query example, the label is predicted with its nearest neighbor among all the class representations.

\textbullet\enspace \textbf{DyFewShot} \cite{DBLPGidarisK18}. 
We introduced this baseline in Section \ref{sec:relatedwork}. 
For this baseline, we extend this baseline to address multi-round few-shot classes: for the present round $C_n^t$, all the preceding classes, including that in $C_b$ and \{$C_n^1\cdots, C_n^{t-1}$\}, are viewed as ``base classes''.

\begin{table*}[t]
\setlength{\tabcolsep}{4.5pt}
\small
  \centering
  \begin{tabular}{ll|c|c|c|c|c|c|c}
 & & $C_b$ & $C_n^1$ & $C_n^2$ & $C_n^3$ & $C_n^4$ & $C_n^5$ & $C_o$\\\hline\hline 
\multirow{5}{*}{$C_b$} & ProtoNet & 87.25$\pm$0.10 & & & & & &53.4$\pm$10.68\\
 & DyFewShot & 81.04$\pm$1.91  & & & & & & 55.01$\pm$2.52 \\
  &  DNNC & 95.96$\pm$0.68& & & & & & 61.89$\pm$4.78 \\
    &  \modelname & \textbf{96.42$\pm$0.41} & & & & & & \textbf{64.73$\pm$3.84}\\
  &  \hybrid &96.12$\pm$0.12 & & & & & &58.92$\pm$1.22 \\\hline
  
 \multirow{5}{*}{$C_n^1$} & ProtoNet & 85.83$\pm$1.94 &31.67$\pm$1.48 & & & & & 43.66$\pm$3.08\\
    & DyFewShot & 81.29$\pm$1.56  &  00.00$\pm$0.00 & & & & & 39.33$\pm$1.25 \\
  &  DNNC & \textbf{95.75$\pm$0.41}& 74.83$\pm$1.64& & & & &\textbf{64.54$\pm$2.02}\\
   &  \modelname & 94.42$\pm$0.21 &75.42$\pm$1.56 & & & & & 56.38$\pm$5.29\\
  &  \hybrid & 95.62$\pm$1.00& \textbf{77.75$\pm$0.25}& & & & & 58.41$\pm$5.10\\\hline
  
 \multirow{5}{*}{$C_n^2$} & ProtoNet & 83.92$\pm$0.33 &24.92$\pm$5.54 &38.83$\pm$3.43 & & & & 31.14$\pm$9.83\\
   & DyFewShot & 81.29$\pm$1.56 & 00.00$\pm$0.00 & 00.50$\pm$0.71 & & & & 33.94$\pm$1.42 \\
 &  DNNC  &95.42$\pm$0.62 &72.92$\pm$4.37 & 75.08$\pm$3.30& & & & \textbf{49.02$\pm$3.23}\\
    &  \modelname & 94.29$\pm$0.16 &71.92$\pm$1.45 &\textbf{84.83$\pm$1.33} & & & & 48.12$\pm$3.20\\
  &  \hybrid &\textbf{96.44$\pm$0.19} &\textbf{76.75$\pm$2.75}& 75.00$\pm$1.00& & & & 42.11$\pm$0.30\\\hline
  
 \multirow{5}{*}{$C_n^3$} & ProtoNet & 81.08$\pm$2.06 &24.33$\pm$5.54 &30.67$\pm$6.17 &22.50$\pm$1.34 & & & 23.62$\pm$6.99\\
   & DyFewShot & 81.29$\pm$1.56 & 00.00$\pm$0.00 & 00.50$\pm$0.71 & 00.00$\pm$0.00 & &  &27.48$\pm$1.24 \\
     &  DNNC &\textbf{95.67$\pm$0.33} & 68.17$\pm$2.37&66.33$\pm$5.02 &71.25$\pm$3.78 & & & \textbf{45.69$\pm$1.73} \\
 & \modelname & 92.71$\pm$0.41& 70.75$\pm$0.54&\textbf{82.83$\pm$2.16} &\textbf{73.92$\pm$2.52} & & &  29.34$\pm$3.31\\
  &  \hybrid & 95.44$\pm$0.44& \textbf{73.62$\pm$0.62}&71.62$\pm$2.62 &73.50$\pm$0.75 & & & 33.69$\pm$3.66 \\\hline

 \multirow{5}{*}{$C_n^4$} & ProtoNet & 81.17$\pm$2.52 &17.83$\pm$2.58 &31.75$\pm$0.94 &24.92$\pm$1.90 &22.25$\pm$3.19 & & 28.19$\pm$4.78\\
 & DyFewShot &81.54$\pm$1.71 &00.25$\pm$0.35&00.17$\pm$0.24&00.00$\pm$0.00&00.00$\pm$0.00& & 23.52$\pm$1.51 \\
  &  DNNC & 95.29$\pm$0.16&68.75$\pm$2.35 &66.75$\pm$3.82 & 67.00$\pm$3.40& 57.75$\pm$1.41& &  42.09$\pm$3.72\\
 & \modelname &91.67$\pm$0.36 &65.92$\pm$2.18 & \textbf{79.92$\pm$1.78}&\textbf{73.75$\pm$0.74} &69.08$\pm$0.12 & & \textbf{45.73$\pm$2.80}\\
  &  \hybrid & \textbf{95.69$\pm$0.06}& \textbf{72.12$\pm$0.62}& 67.75$\pm$1.25&70.25$\pm$0.25 & \textbf{72.62$\pm$1.38} & &  38.85$\pm$0.89\\\hline
  
 \multirow{5}{*}{$C_n^5$} & ProtoNet & 80.00$\pm$2.65 &21.83$\pm$5.45 &29.17$\pm$3.70 & 24.67$\pm$3.12& 23.17$\pm$3.60& 30.33$\pm$4.17& 29.24$\pm$2.96\\
   & DyFewShot &81.50$\pm$1.27 &00.08$\pm$0.12&00.83$\pm$0.62&00.00$\pm$0.00&00.00$\pm$0.00&00.50$\pm$0.71& 21.23$\pm$1.34\\
  &  DNNC  &95.12$\pm$0.47 &67.50$\pm$0.89 &67.92$\pm$4.70 & 64.42$\pm$4.17& 52.42$\pm$1.20& 53.33$\pm$2.09& 30.46$\pm$5.92\\
 & \modelname &89.17$\pm$0.60 & 65.08$\pm$2.45&\textbf{78.50$\pm$0.94} & \textbf{69.08$\pm$1.12}&\textbf{68.25$\pm$0.35} &\textbf{70.67$\pm$1.30}& \textbf{39.48$\pm$1.45}\\
  &  \hybrid & \textbf{95.56$\pm$0.06}&\textbf{68.75$\pm$2.75} & 67.38$\pm$0.62& 63.75$\pm$1.75&65.12$\pm$3.62 &61.62$\pm$2.38 &37.65$\pm$0.44\\ \hline
\end{tabular}
\caption{System performance with base classes on the benchmark \intentdata.
}\label{tab:intentresults}
\end{table*}

\begin{figure}[t]
\centering
\subfigure[\vspace{-0.1in}
\intentdata.]{\includegraphics[width=3.8cm]{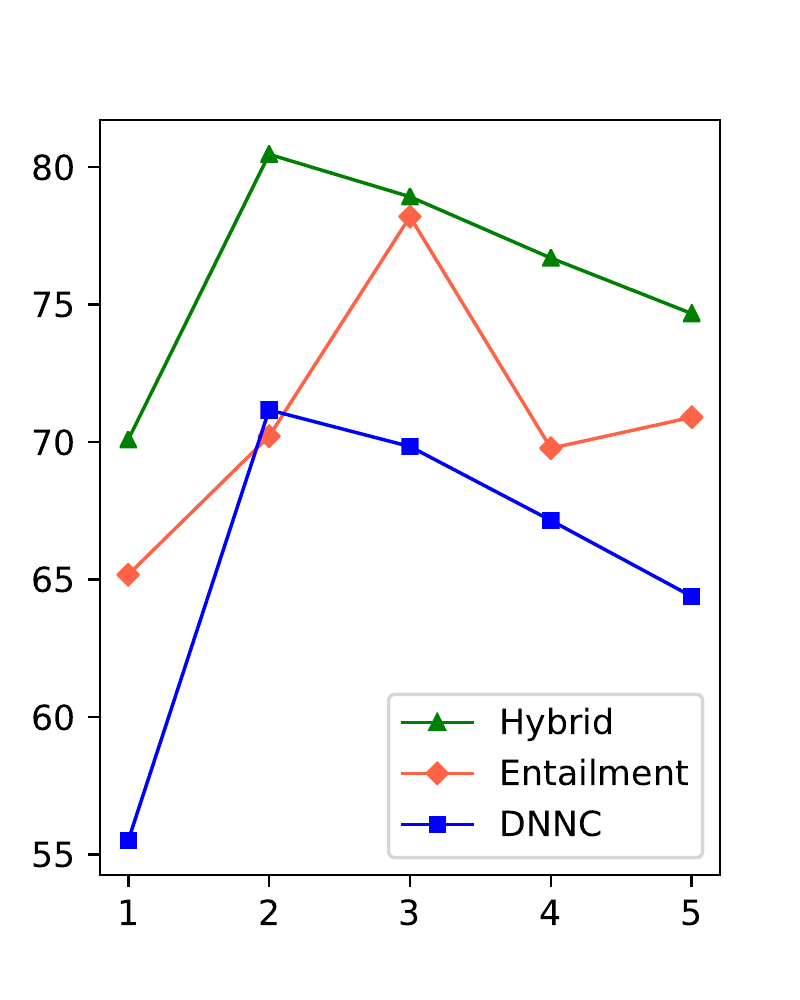}}
\vspace{-0.08in}
\subfigure[\vspace{-0.1in}
\relationdata.]{\includegraphics[width=3.8cm]{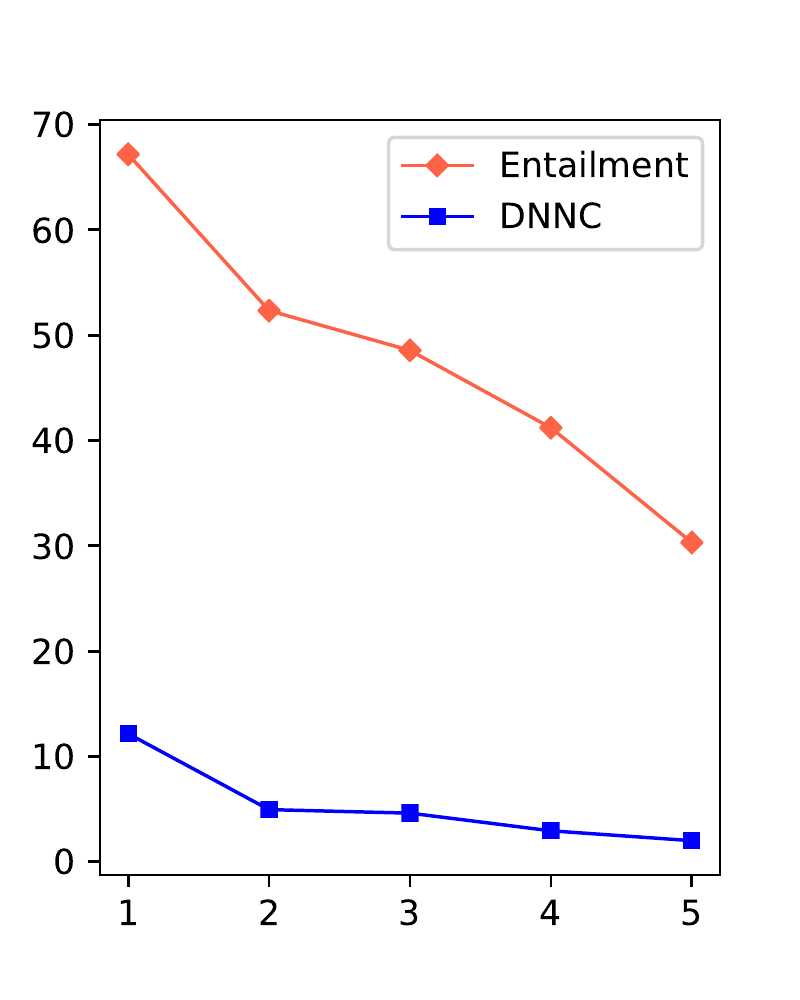}}
\vspace{-0.08in}
\caption{Average performance on new classes in different rounds. The x axis is the number of round and y is the average accuracy on new classes in this round.}
\vspace{-0.1in}
\label{fig:performance_withoutbase}
\end{figure}

\paragraph{Implementation and setting.}
For DNNC, \modelname, and \hybrid , we use the MNLI \cite{williams2018broad} dataset to pre-train these models. 
All systems are implemented through the Huggingface Transformers package.
For both pre-training and fine-tuning, we set the learning rate as 1e-6, the batch size is 16. We run 20 epochs for the pre-training. For the fine-tuning process, we run 5 epochs on \intentdata\enspace and 50 epochs on \relationdata.
We run the same program with 3 different seeds and report the average performance. Accuracy is reported for \{$C_b$, $C_n^1$, $\cdots$, $C_n^5$\} and F1 score for $C_o$.

\subsection{Experimental results}
As the problem formulation presented in Section \ref{sec:probdefinitino}, we want to investigate two questions. $\mathcal{Q}_1$: can our system get better performance on each round? $\mathcal{Q}_2$: can our system hold more stable performance during the incremental learning process? We answer these questions separately under the incremental learning setting with or without base classes.

\begin{figure}[t]
\centering
\subfigure[Average Performance.]{\includegraphics[width=3.8cm]  {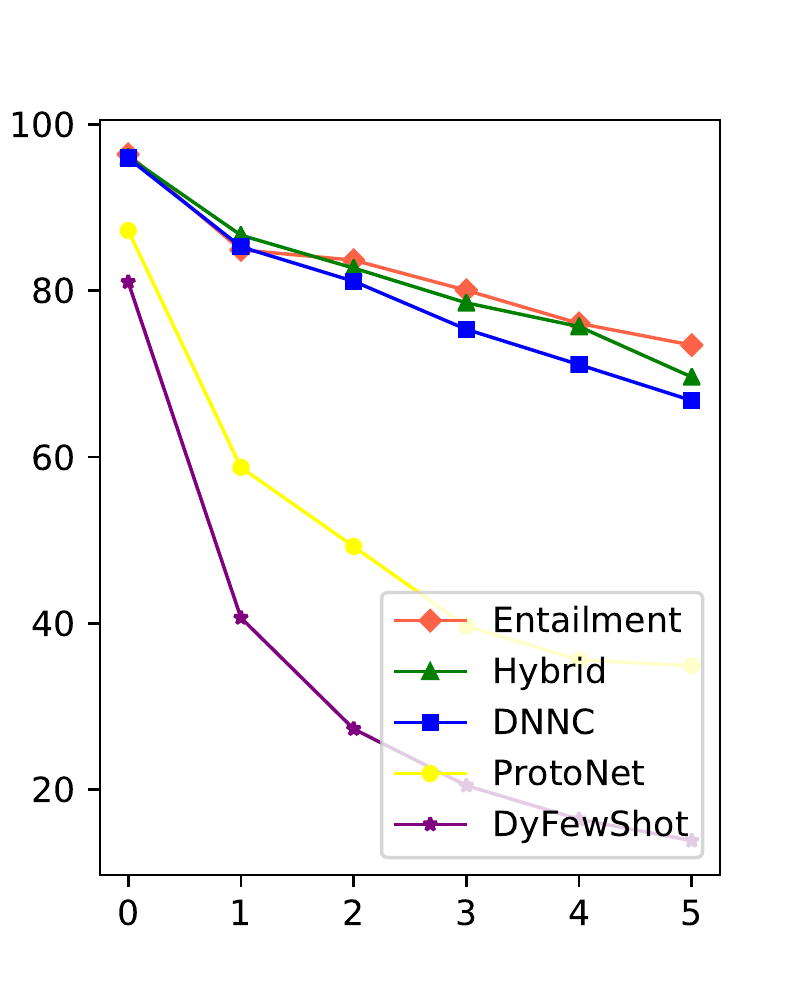}} \vspace{-0.05in}
\subfigure[Performance drop rate.]{\includegraphics[width=3.8cm]{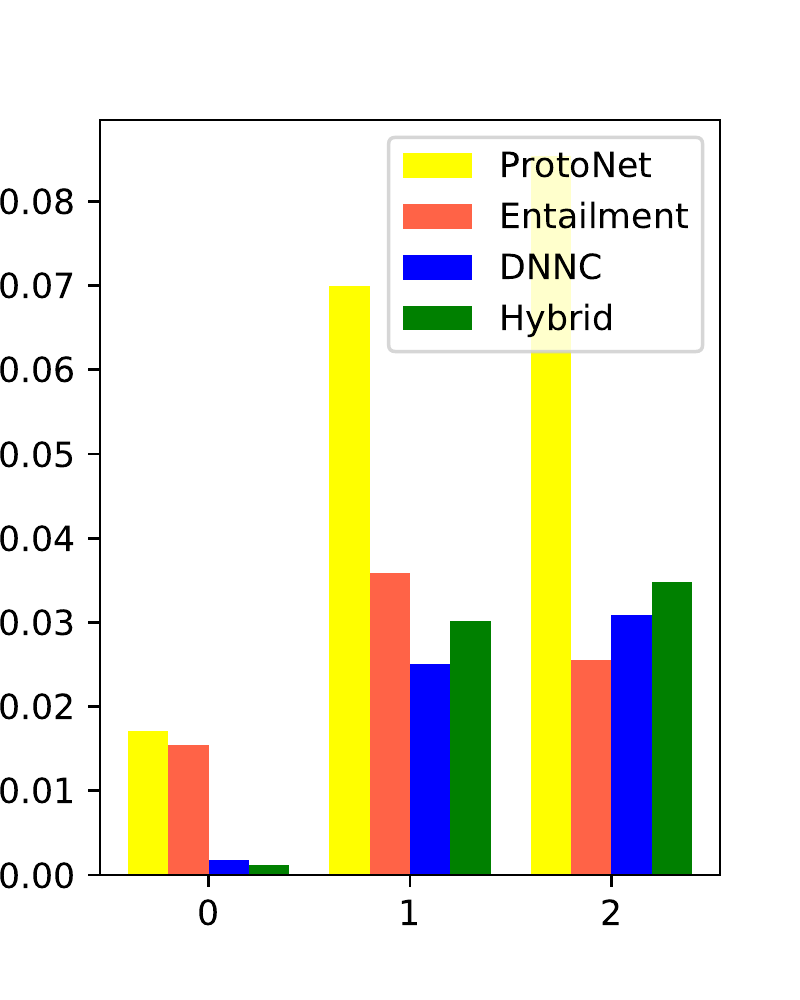}} \vspace{-0.05in}
\caption{Performance analysis on \intentdata\enspace with base classes. In Figure (a), X axis is the number of round, where 0 indicates the round of base classes; Y axis is the average performance on all the seen classes in this round (including base classes $C_b$ and new classes $C_n^1$, ..., $C_n^i$). In Figure (b), X axis indicates a subset of classes, where 0 indicates base classes and 1 indicates new classes in $C_n^1$; Y axis is the performance drop rate $d$ for different subsets.}
\label{fig:performance_with_base}
\vspace{-0.05in}
\end{figure}

\paragraph{Incremental learning without base classes.}
Tables \ref{tab:intentresults_withoutbase}$\sim$\ref{tab:relationresults_withoutbase} list the results on two benchmarks, \intentdata\enspace and \relationdata\enspace,  for the setting without base classes, respectively.  For the \intentdata\enspace benchmark, we compare \modelname\enspace with DNNC, together with their hybrid model \hybrid\enspace for 5 rounds. For the \relationdata\enspace benchmark, we only compare \modelname\enspace with DNNC since \hybrid\enspace is not applicable for this dataset. The label (relation type) is not compatible with the input instance (an utterance with an entity pair). Therefore, we can not mix the pairs from these two models (\modelname\enspace and DNNC) to train a hybrid model.


As for question $\mathcal{Q}_1$, we find that \modelname\enspace and \hybrid\enspace outperform all the baselines. These results show the effectiveness of formalizing text classification as a textual entailment problem.
For the benchmark \intentdata\enspace, the hybrid model, \hybrid, achieves the best performance since it has the largest number of entailment pairs (by combining pairs from two models) for the training. It shows in the extreme case that no base classes are available, the more data the better. For the benchmark \relationdata, this task is much more difficult compared to intent detection due to the complicity of the training examples (utterances with entity pairs). DNNC does not perform well for this task since comparing two complex examples can not benefit from the pre-training entailment model.

As for $\mathcal{Q}_2$, we show the average performance change on new classes in Figure \ref{fig:performance_withoutbase}. For \intentdata, the average performance of new classes increases in the beginning then drops for the remaining rounds. This might due to the lack of training data in the first found. For \relationdata, the average performance drops dramatically due to this task is much more difficult.

\paragraph{Incremental learning with base classes.}
The results on \intentdata\enspace with base classes are shown in Table \ref{tab:intentresults}. 
We compare our systems \modelname\enspace and \hybrid\enspace with three baselines: DNNC, ProtoNet, and DyFewShot. This setting is evaluated incrementally on base classes, five rounds of new classes, and OOD.


As for question $\mathcal{Q}_1$, we summarize our observations as follows. 
(i) Pre-trained models (\modelname, \hybrid, and DNNC) work much better than few-shot learning methods (ProtoNet and DyFewShot) which means pre-training from a large-scale entailment dataset helps a lot in this setting. (ii) Our proposed models, \modelname\enspace and \hybrid\enspace obtain comparable performances and they outperform all the other baselines consistently in all test classes for the whole timeline. This shows the effectiveness of our proposed method of generating (utterance, label) entailment pairs.

To answer $\mathcal{Q}_2$ in this setting, we propose a new evaluation metric, performance drop rate $d$, to evaluate the performance change along the timeline, i.e., how fast the performance drops when adding new rounds of classes into the system. For example, the performances on base classes decrease when incrementally adding five rounds of new classes into the system. Given a list of performance results for a certain subset of classes (for example, base classes) on $m$ rounds, $r = (r_1, r_2, ..., r_m)$, we calculate the performance drop rate as the average drop rate of different rounds $d =\frac{1}{m-1}  \sum_{i=0}^{m-1} (r_i - r_{i+1}) / r_i $. In the experiments, we calculate $d$ for four methods on base classes, new classes in round1 and round2 separately. The average drop rate of DyFewShot is not reported since there are 0.0 values in the performance. 

In Figure \ref{fig:performance_with_base}, we show the average performance on all the seen classes in different rounds (including base classes and all the seen new classes) in (a) and the performance drop rate $d$ in (b). As shown in Figure \ref{fig:performance_with_base} (a), the average performance drops with the increase of round numbers. We can also observe that our proposed models \modelname\enspace and \hybrid\enspace achieve the best performance on the average performance on all the seen classes, including base classes and new classes.
Figure \ref{fig:performance_with_base} (b) shows the the performance drop rate $d$ for different models. ProtoNet and \modelname\enspace have higher drop rate than DNNC and \hybrid\enspace on base classes.
For new classes in round1 and round2, the drop rate on ProtoNet is much higher than all the entailment methods. In summary, \modelname\enspace achieves the best performance on the average accuracy of all seen classes, while DNNC is more stable and has a lower performance drop rate. \hybrid\enspace combines the advantages of both models by combining these two models together.



\section{Conclusion}
In this work, we define a new challenge in the NLP domain, incremental few-shot text classification with multi-round new classes in two settings: with or without base classes. In addition to the problem formulation, we also release two benchmark datasets for this particular challenge: \intentdata\enspace and \relationdata. Two approaches, \modelname\enspace and \hybrid\enspace are proposed to solve this problem. They convert the text classification problem into textual entailment and make the maximum utilization of the pre-training textual entailment model. Extensive experiments are conducted and the results consistently show the effectiveness of our proposed models.

\section*{Acknowledgments}
We thank the reviewers for their valuable comments.
This work is supported in part by NSF under grants III-1763325, III-1909323, and SaTC-1930941.

\bibliography{naacl2021}
\bibliographystyle{acl_natbib}

\end{document}